\documentclass[11pt]{article}

\usepackage{acl2000}
\usepackage{macros}
\usepackage{multicol}  
\usepackage{amssymb}   
\usepackage{isolatin1} 

\include{epsf}

\title{Mapping WordNets Using Structural Information}
\author{J. Daud\'e, L. Padr\'o \& G. Rigau\\
        TALP Research Center\\
        Departament de Llenguatges i Sistemes Inform\`atics\\
        Universitat Polit\`ecnica de Catalunya. Barcelona\\
        {\tt \{daude,padro,g.rigau\}@lsi.upc.es} }

\begin{document}
\maketitle

\begin{abstract}
We present a robust approach for linking already existing lexical/semantic
hierarchies. We used a constraint satisfaction algorithm (relaxation
labeling) to select --among  a set of candidates-- the node in a 
target taxonomy that bests matches each node in a source taxonomy.
In particular, we use it to map the nominal part of WordNet 1.5 onto
WordNet 1.6, with a very high precision and a very low remaining ambiguity.
\end{abstract}

\section{Introduction}

There is an increasing need of having available general, accurate and
broad coverage multilingual lexical/semantic resources for developing 
{\sc nl} applications. Thus, 
a very active field inside {\sc nl} during the last years has been the 
fast development of generic language resources.

Several attempts have been performed to connect already existing ontologies. 
In \cite{ageno94}, a Spanish/English bilingual dictionary is used to
(semi)automatically link Spanish and English taxonomies extracted 
from {\sc dgile} \cite{dgile87} and {\sc ldoce} \cite{ldoce87}.
Similarly, a simple automatic approach for linking 
Spanish taxonomies extracted from {\sc dgile} to WordNet \cite{miller91} 
synsets is proposed in \cite{rigau95}. 
The work reported in \cite{knight94} focuses on the construction of Sensus, a 
large knowledge base for supporting the Pangloss machine translation system. 
In \cite{okumura94} (semi)automatic 
methods for associating a Japanese lexicon to an English ontology using a bilingual 
dictionary are described. Several experiments aligning {\sc edr} and 
WordNet ontologies are described in \cite{utiyama97}. Several lexical resources and 
techniques are combined in \cite{atserias97} to map Spanish words from a 
bilingual dictionary to WordNet, and in \cite{farreres98} the use of the 
taxonomic structure derived from a monolingual {\sc mrd} is proposed as an aid 
to this mapping process.

  The use of relaxation labeling algorithm to attach substantial fragments
of the Spanish taxonomy derived from {\sc dgile} \cite{rigau98} to the English WordNet 
using a bilingual dictionary for connecting both hierarchies, has been
reported in \cite{Daude99b}.

  In this paper we use the same technique to map {\sc wn1.5} to {\sc wn1.6}. The 
aim of the experiment is twofold: First, show that the method is
general enough to link any pair of ontologies. Second, evaluate our 
taxonomy linking procedure, by comparing our results with other
{\sc wn1.5} to {\sc wn1.6} existing mappings.

 This paper is organized as follows: In section \ref{application} we describe the 
used technique (the relaxation labeling algorithm) and its application to hierarchy
mapping. In section \ref{constraints} we describe the constraints used in the relaxation
process, and finally, after presenting some experiments and results, we 
offer some conclusions and outline further lines of research.

\section{Application of Relaxation Labeling to {\sc nlp}}
\label{application}

Relaxation labeling ({\sc rl}) is a generic name for a family of iterative
algorithms which perform function optimization, based on local
information.  See \cite{Torras89} for a summary.
Its most remarkable feature is that it can deal with any kind of constraints,
thus, the model can be improved by adding any constraints available, and
the algorithm is independent of the complexity of the model. That is,
we can use more sophisticated constraints without changing the
algorithm. 

The algorithm has been applied to {\sc pos} tagging 
\cite{Marquez97a}, shallow parsing \cite{Voutilainen97} and
to word sense disambiguation \cite{Padro98a}.

Although other function optimization algorithms 
could have been used (e.g. genetic algorithms, 
simulated annealing, etc.), we found {\sc rl} to be 
suitable to our purposes, given its ability to use models
based on context constraints, and the existence of previous
work on applying it to {\sc nlp} tasks.

Detailed explanation of the algorithm can be found in \cite{Torras89},
while its application to {\sc nlp} tasks, advantages and drawbacks are
addressed in \cite{Padro98a}.

\subsection{Algorithm Description}

The Relaxation Labeling algorithm deals with a set of variables (which may
represent words, synsets, etc.), each of which may take one among several 
different labels ({\sc pos} tags, senses, {\sc mrd} entries, etc.). 
There is also 
a set of constraints which state compatibility or incompatibility of a 
combination of pairs variable--label.

 The aim of the algorithm is to find a weight assignment for each
possible label for each variable, such that (a) the weights for 
the labels of the same variable add up to one, and (b) the weight
assignment satisfies --to the maximum possible extent-- the 
set of constraints. 

 Summarizing, the algorithm performs constraint satisfaction to solve
a consistent labeling problem. The followed steps are:

\begin{enumerate}
\item Start with a random weight assignment.
\item Compute the {\sl support} value for each label of each variable. 
       Support is computed according to the constraint set and to the
       current weights for labels belonging to context variables.
\item Increase the weights of the labels more compatible with the
      context (larger support) and decrease those of
      the less compatible labels (smaller support). Weights are changed
      proportionally to the support received from the context. 
\item If a stopping/convergence criterion is satisfied,
      stop, otherwise go to step 2. We use the
      criterion of stopping when there are no more changes, although
      more sophisticated heuristic procedures may also be used to stop
      relaxation processes \cite{Eklundh78,Richards81}. 
\end{enumerate}

The cost of the algorithm is proportional to the product of the number of 
variables by the number of constraints.

\subsection{Application to taxonomy mapping}

   As described in previous sections, the problem we are dealing
  with is to map two taxonomies. In this particular case, we are
  interested in mapping {\sc wn1.5} to {\sc wn1.6}, that is, assign 
  each synset of the former to at least one synset of the later.

  The modeling of the problem is the following:
   \begin{itemize}
      \item Each {\sc wn1.5} synset is a variable for the relaxation
           algorithm. We will refer to it as {\sl source synset} and
           to {\sc wn1.5} as {\sl source taxonomy}.
      \item The possible labels for that variable are all the {\sc
           wn1.6} synsets which contain a word belonging to the source
           synset. We will refer to them as {\sl target synsets} and
           to {\sc wn1.6} as {\sl target taxonomy}.
      \item The algorithm will need constraints stating whether a {\sc
          wn1.6} synset is 
           a suitable assignment for a {\sc wn1.5} synset. As described
           in section~\ref{constraints}, these constraints will
           rely on the taxonomy structure. 
   \end{itemize}

\section{The Constraints}
\label{constraints}

   Constraints are used by the relaxation labeling algorithm to increase or decrease the
weight for a variable label. In our case, constraints increase the
weights for the connections between a source synset and a
target synset. Increasing the weight for a connection implies decreasing
the weights for all the other possible connections for the same source
synset.
 To increase the weight for a connection, constraints look for already connected 
nodes that have the same relationships in both taxonomies.

   Although there is a wide range of relationships between WordNet synsets which 
can be used to build constraints, we have focused on the hyper/hyponym relationships.
That is, we increase the weight for a connection when the involved nodes have
hypernyms/hyponyms also connected. We consider hyper/hyponym relationships 
either directly or indirectly (i.e. ancestors or descendants), depending on
the kind of constraint used.

   Figure~\ref{f-exemple1} shows an example of possible connections between two
taxonomies. Connection $C_4$ will have its weight increased due to $C_5$, $C_6$ and
$C_1$, while connections $C_2$ and $C_3$ will have their weights decreased.

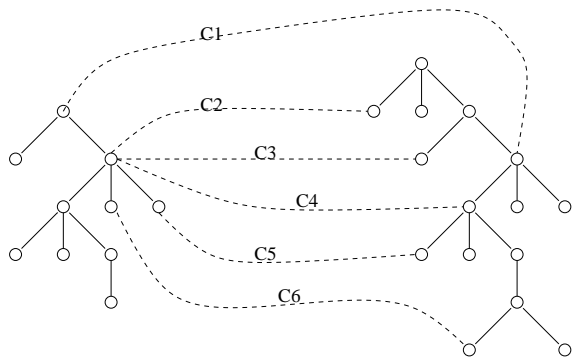
\figeps{exemple1}{Example of connections between taxonomies.}{f-exemple1}

   We distinguish different kinds of constraints, depending on whether we consider
hyponyms, hypernyms or both, on whether we consider those relationships direct or
indirect, and on in which of both taxonomies we do so.
   Each constraint may be used alone or combined with others. 

  Below we describe all kinds of constraint used. They are labeled with a three--character 
code ({\sc xyz}), which must be read as follows: The first character ({\sc x}) indicates how the
hyper/hyponym relationship is considered in the source taxonomy: only 
for {\sl immediate} nodes ({\sc i}) or for {\sl any} ({\sc a}) ancestor/descendant.
The second character ({\sc y}) codes the same information for the target taxonomy
side. The third character indicates whether the constraints requires the
existence of a connected hypernym ({\sc e}), hyponym ({\sc o}), or both ({\sc b}).

\begin{description}
\item[{\sc iie} constraint:] 
   The simplest constraint is to check whether the connected nodes
 have respective direct hypernyms also connected. {\sc iie} stands
 for immediate source({\sc i}), immediate target ({\sc i}) 
 hypernym ({\sc e}). 
 
   This constraint will increase the weights for those connections in
 which the immediate hypernym of the source node is connected
 with the immediate hypernym of the target node. 

\item[{\sc iio} constraint:]
 This constraint increases the weight for that connections in which an
immediate hyponym of the source node is connected
to an immediate hyponym of the target node.

\item[{\sc iib} constraint:]
 This constraint increases the weight for the connections in which the
immediate hypernym of the source node is connected
to the immediate hypernym of the target node and an
immediate hyponym of the source is connected
to an immediate hyponym of the target.

\item[{\sc ii} constraints.]
If we use constraints {\sc iie}, {\sc iio} and {\sc iib} at the same time, 
weights will be modified
for words matching any of the constraints. That is, we are additively
combining constraints. In the case where two of them apply, their
effects will be added. If they have opposite effects, they will 
cancel each other.  Figure~\ref{f-ii} shows a graphical representation of 
all {\sc ii} constraints.
\end{description}

\figeps{ii}{{\sc ii} constraints.}{f-ii}

  The arrows indicate an immediate hypernymy relationship. The nodes on the
left hand side correspond to the source taxonomy and the nodes on the right
to the target hierarchy. The dotted line is the connection which weight will be    
increased due to the existence of the connection indicated with a continuous line.

\begin{description}
\item[{\sc aie} constraint:]

 This constraint increases the weight for the connections in which an
ancestor of the source node is connected
to the immediate hypernym of the target node. 

\item[{\sc aio} constraint:]
 This constraint increases the weight for the connections in which a
descendant of the source node is connected
to an immediate hyponym of the target node.

\item[{\sc aib} constraint:]
 This constraint increases the weight for the connections in which 
an ancestor of the source node is connected
to the immediate hypernym of the target node and
a descendant of the source node is connected
to an immediate hyponym of the target node. 

\item[{\sc ai} constraints.]
  If we use constraints {\sc aie}, {\sc aio} and {\sc aib} simultaneously, 
we apply either a hypernym constraint, either a hyponym constraint or either
both of them. In the last case, the joint constraint is also applied. This means
than connections with matching hypernym and hyponym will have their weights doubly 
increased. Figure~\ref{f-ai} shows a graphical representation of 
all {\sc ai} constraints.
\end{description}

\figeps{ai}{{\sc ai} constraints.}{f-ai}

  In this figure, the $+$ sign indicates that the hypernymy relationship 
represented by the arrow does not need to be immediate. In this case, this 
iteration is only allowed in the source taxonomy.

\begin{description}
\item[{\sc ia} constraints:] Are symmetrical to {\sc ai} constraints. In
this case, recursion is allowed only on the target taxonomy.
\end{description}

 Figure~\ref{f-ia} shows a graphical representation of 
all {\sc ia} constraints.

\figeps{ia}{{\sc ia} constraints.}{f-ia}

\begin{description}
\item[{\sc aa} constraints:] Include the above combinations, but allowing
recursion on both sides.
\end{description}

 Figure~\ref{f-aa} shows a graphical representation of 
all {\sc aa} constraints.

\figeps{aa}{{\sc aa} constraints.}{f-aa}

\section{Experiments and Results}
\label{experiments}

  In the performed tests we used simultaneously all 
constraints with the same recursion pattern. This yields the packs: {\sc ii},
{\sc ai}, {\sc ia} and {\sc aa}. Results are reported only for the
later, since it is the most informed constraint set.

  We also compared our mapping with the {\sl SenseMap} provided by 
Princeton\footnote{See {\sc wn} web page at \\
                   http://www.cogsci.princeton.edu/\ona wn/},
and the coincidence was quite high, specially in the cases in which 
{\sl SenseMap} has a high confidence score. Details can be found in
section~\ref{sec-coinc}.

  In order to perform the comparison, we had to convert {\sl SenseMap},
which is a sense mapping (that is, it maps each {\sl variant} in {\sc wn}1.5
to a variant in {\sc wn}1.6), into a {\sl synset} mapping, which is what
our algorithm does. Since synsets are coarser than senses, the
conversion is straightforward. When two senses in the same 1.5 synset
were assigned two senses in different 1.6 synsets, we took both
targets as valid, slightly increasing the remaining ambiguity
of {\sl SenseMap}. 

The results are computed over the synsets with at least
one candidate connection, which represent 99.1\% of {\sc wn1.5}.
We consider {\sl ambiguous} synsets those with more than one 
candidate connection.

  Table~\ref{taula-cover} presents the amount of nodes for which
disambiguation is performed, and some candidate connections 
discarded (i.e. they do not keep as possible {\sl all} the candidates).

\begin{table}[htb] \centering
\begin{tabular}{l|c|c|}
               & ambiguous & overall  \\ \hline
{\sl SenseMap} & 98.0\%    & 99.2\% \\
{\sc rl}       & 99.8\%    & 99.9\% \\ \hline
\end{tabular}
\caption{Coverage of {\sc wn1.5} for both mappings.}
\label{taula-cover}
\end{table}

Table~\ref{taula-prec}
presents an estimation of how many of those
assignment were right, as well as the precision for {\sl SenseMap},
computed under the same conditions. Those figures were computed by manually
linking to {\sc wn1.6} a sample of 1900 synsets randomly chosen from {\sc wn}1.5, and
then use this sample mapping as a reference to evaluate all mappings.
 These figures show that our system performs a better
mapping than {\sl SenseMap}. The difference between both
mappings is significant at a 95\% confidence level.

\begin{table}[htb] \centering
\begin{tabular}{l|c|c|}
                      & ambiguous      & overall        \\ \hline
{\sl SenseMap}        & 93.3\%--96.9\% & 96.9\%--98.6\% \\ \hline
{\sc rl} ($\delta=0.3$) & 96.5\%--97.7\% & 98.4\%--98.9\% \\
{\sc rl} ($\delta=0.4$) & 97.0\%--97.6\% & 98.6\%--98.9\% \\
{\sc rl} ($\delta=0.5$) & 97.2\%--97.6\% & 98.7\%--98.9\% \\ \hline
\end{tabular}
\caption{Precision--recall results for both {\sc wn}1.5--{\sc wn}1.6 mappings.}
\label{taula-prec}
\end{table}

Since relaxation labeling performs a weight assignment for each possible
connection, we can control the remaining ambiguity (and thus the 
recall/precision tradeoff) by changing the threshold ($\delta$)
that the weight for a connection has to reach to be considered a
solution. Although higher thresholds maintain recall and
produce a higher precision, differences are not statistically 
significant.

\subsection{Coincidence of Both Mappings}
\label{sec-coinc}

   For each confidence group in the Princeton mapping,
   the {\sl soft agreement} column in table~\ref{taula-up-agree} 
   indicates the
   percentage of {\sc wn}1.5 synsets in which our system proposes at least one
   connection also proposed by the Princeton mapping. The {\sl hard agreement} column
   indicates the amount of connections proposed by our system also
   proposed by Princeton mapping. 

\begin{table*}[htb] \centering
\begin{tabular}{|c|cc|cc|cc|}
                  & \multicolumn{6}{c|}{Agreement} \\ \cline{2-7}
       confidence & \multicolumn{2}{c|}{$\delta=0.3$} & \multicolumn{2}{c|}{$\delta=0.4$} & \multicolumn{2}{c|}{$\delta=0.5$}   \\ 
       group      &{\sl hard} &{\sl soft} &{\sl hard}&{\sl soft}&{\sl hard}&{\sl soft}  \\ \hline 
       monosemous & 96.9\% & 97.3\% & 97.0\% & 97.3\% & 97.1\% & 97.2\%\\ \hline
         100\%    & 88.6\% & 90.4\% & 89.1\% & 90.1\% & 89.5\% & 89.8\%\\ 
          90\%    & 87.9\% & 89.8\% & 88.4\% & 89.3\% & 88.7\% & 89.1\%\\ 
          80\%    & 69.3\% & 70.2\% & 70.1\% & 70.5\% & 70.4\% & 70.4\%\\ 
          70\%    & 76.5\% & 78.0\% & 76.4\% & 77.6\% & 76.5\% & 76.8\%\\ 
          60\%    & 53.8\% & 53.8\% & 53.8\% & 53.8\% & 53.8\% & 53.8\%\\ 
          50\%    & 68.4\% & 81.2\% & 72.7\% & 77.4\% & 72.7\% & 77.4\%\\ 
          40\%    & 50.7\% & 50.8\% & 50.8\% & 50.8\% & 50.8\% & 50.8\%\\ 
          30\%    & 65.3\% & 65.3\% & 65.3\% & 65.3\% & 65.3\% & 65.3\%\\ 
          20\%    & 32.6\% & 32.6\% & 32.6\% & 32.6\% & 32.6\% & 32.6\%\\ 
       subtotal   & 87.3\% & 89.1\% & 87.8\% & 88.8\% & 88.3\% & 88.6\%\\ \hline\hline
         Total    & 93.6\% & 94.5\% & 93.8\% & 94.4\% & 94.1\% & 94.2\%\\ \hline
\end{tabular}
\caption{Agreement between both mappings.}
\label{taula-up-agree}
\end{table*}
 
      The agreement between both systems is quite high, specially for the groups 
   with a high confidence level. This is quite reasonable, since a perfect system
   would be expected to agree with the assignments in 20\% confidence group of
   {\sl SenseMap} only about 20\% of the times. It also must be taken into account 
   that for low confidence groups, {\sl SenseMap} is much more ambiguous.

     The average remaining ambiguity in Princeton mapping and in 
   the mapping performed by the relaxation labeling algorithm is 
   shown respectively in columns {\sl SenseMap ambiguity} and 
   {\sl {\sc rl} ambiguity} of table~\ref{taula-up-amb}.

\begin{table*}[htb] \centering
\begin{tabular}{|cc|c|ccc|}
       Confidence   &       &{\sl SenseMap} &  \multicolumn{3}{c|}{{\sc rl} ambiguity} \\ 
       group        & Size  & ambiguity  & $\delta=0.3$ & $\delta=0.4$ & $\delta=0.5$ \\ \hline
       monosemous   & 45807 & 1.003   &  1.001  & 1.001 & 1.001 \\ \hline
         100\%      & 20075 & 1.000   &  1.020  & 1.011 & 1.003 \\ 
          90\%      &  2977 & 1.007   &  1.022  & 1.010 & 1.005 \\ 
          80\%      &   326 & 1.080   &  1.018  & 1.009 & 1.000 \\ 
          70\%      &   249 & 1.024   &  1.020  & 1.012 & 1.004 \\ 
          60\%      &    93 & 1.043   &  1.000  & 1.000 & 1.000 \\ 
          50\%      &    32 & 1.063   &  1.125  & 1.064 & 1.064 \\ 
          40\%      &    67 & 1.448   &  1.031  & 1.015 & 1.000 \\ 
          30\%      &    65 & 1.569   &  1.000  & 1.000 & 1.000 \\ 
          20\%      &   209 & 2.215   &  1.031  & 1.025 & 1.020 \\
       subtotal     & 24093 & 1.016   &  1.020  & 1.011 & 1.003 \\ \hline\hline
         Total      & 69900 & 1.007   &  1.007  & 1.006 & 1.001 \\ \hline
\end{tabular}
\caption{Average remaining ambiguity of both mappings.}
\label{taula-up-amb}
\end{table*}

   Our system proposes, in most cases, a unique {\sc wn}1.6 synset
 for each {\sc wn}1.5 synset. The average ranges from 1.001 to 1.007 
 proposals per synset depending on the chosen $\delta$ threshold,
 while the Princeton mapping has an average of 1.007.

   Summarizing, the obtained results point that our system is able to
 produce a less ambiguous assignment than {\sl SenseMap}, with a
 significantly higher accuracy and wider coverage. 

   In addition, our system only uses structural information 
(namely, hyper/hyponymy relationships) while {\sl SenseMap} uses
 synset words, glosses, and other information in WordNet. 
   On the one hand, when information other than taxonomy structure is used results might
 be even better. On the other hand, for cases in which such information
 is not available (e.g. further development of EuroWordNets in new languages),
 structure may provide a reliable basis.

\section{Conclusions \& Further Work}

   We have applied the relaxation labeling algorithm to assign an appropriate
node in a target taxonomy to each node in a source taxonomy, using only hyper/hyponymy
information.

 Results on {\sc wn1.5} to {\sc wn1.6} mapping have been reported.
The high precision achieved provides further evidence that this 
technique --previously used in \cite{Daude99b} to link a Spanish 
taxonomy to {\sc wn1.5}-- constitutes
an accurate method to connect taxonomies, either for the same
or different languages. Further extensions of this technique to
include information other than structural may result in a valuable 
tool for those concerned with the development and improvement 
of large lexical or semantic ontologies.

   The results obtained up to now seem to indicate that:

\begin{itemize}
  \item The relaxation labeling algorithm is a good technique to link
     two different hierarchies. For each node with several possible connections,
     the candidate that best matches the surrounding structure is selected.
 \item The structural information provides enough knowledge to
   accurately link taxonomies. Experiments on mapping taxonomies automatically
   extracted from a Spanish {\sc mrd} to {\sc wn}1.5 \cite{Daude99b}
   show that the technique may be useful even when both taxonomies
   belong to different languages or have structures less similar than
   in the case reported in this paper.
 \item The system produces a good assignment for {\sc wn} mapping, based only
     on hyper/hyponymy relationships, which is specially useful when no other 
     information is available (i.e. in the case of mapping the EuroWN hierarchies).
     The remaining ambiguity is low with a high accuracy, and
     precision--recall tradeoff may be controlled by adjusting the $\delta$ threshold.
\end{itemize}
 
   Some issues to address for improving the algorithm performance, and to
 exploit its possibilities are:

\begin{itemize}
  \item Use other relationships than hyper/hyponymy as constraints
      to select the best connection. Relationships as sibling, cousin, etc. could
      be used. In addition, {\sc wn} provides other relationships such as synonymy, 
      meronymy, etc. which could also provide useful constraints.
  \item Use other available information, such as synset words, glosses, etc. in 
     the {\sc wn} to {\sc wn} mapping task. 
  \item Link the verbal, adjectival, and adverbial parts of {\sc wn}1.5 and {\sc wn}1.6.
  \item Test the performance of the technique to link other structures (e.g {\sc wn-edr, wn-ldoce}, 
       dutch-{\sc wn}, italian-{\sc wn}, \ldots).
  \item Use it to link taxonomies for new languages to EuroWordNet.
  \item Give a step beyond the source-to-target vision, and map the
      taxonomies in a symmetrical philosophy, that is, each node of each
      taxonomy is assigned to a node in the other taxonomy. This should
      increase the coverage, and reinforce/discard connections that may be weak 
      when assigned only in one direction. This could even open the
      doors to many-to-many taxonomy mapping.
\end{itemize}

\section{Acknowledgments}

This research has been partially funded by the the UE Commission (NAMIC IST-1999-12392), by the 
Spanish Research Department (TIC98-423-C06-06) and by the Catalan Research Department, through 
the CREL project and the Quality Research Group Programme (GRQ93-3015).

\bibliographystyle{acl}
\bibliography{/usr/usuaris/ia/padro/articles/fullbib}

\end{document}